\documentclass[letterpaper, 10 pt, conference]{ieeeconf}  % Comment this line out if you need a4paper

\IEEEoverridecommandlockouts                              % This command is only needed if 
\title{\LARGE \bf
\textit{eNavi}: Event-based Imitation Policies for Low-Light Indoor \\ Mobile Robot Navigation
}

\author{Prithvi Jai Ramesh$^{1}$, Kaustav Chanda$^{1}$, Krishna Vinod$^{1}$, Joseph Raj Vishal$^{1}$ \\
Yezhou Yang$^{1}$ and Bharatesh Chakravarthi$^{1}$% <-this % stops a space
\thanks{$^{1}$School of Computing and Augmented Intelligence, Arizona State University, Tempe, AZ, USA.
        {\tt\small pjramesh@asu.edu and bshettah@asu.edu}}%
}

\usepackage{graphicx}      % Required for images
\usepackage{amsmath}       % Math symbols
\usepackage{amssymb}       % More math symbols
\usepackage{url}           % Handle URLs
\usepackage{booktabs}      % For professional looking tables
\usepackage{multirow}      % For multirow cells in tables
\usepackage[table]{xcolor} % For colored table rows
\usepackage{cite}          % Fixes citation ordering like 
\usepackage[caption=false,font=footnotesize]{subfig}

\begin{document}

\maketitle
\thispagestyle{empty}
\pagestyle{empty}

\begin{abstract}

Event cameras provide high dynamic range and microsecond-level temporal resolution, making them well-suited for indoor robot navigation, where conventional RGB cameras degrade under fast motion or low-light conditions. Despite advances in event-based perception, spanning detection, SLAM, and pose estimation, there remains limited research on end-to-end control policies that exploit the asynchronous nature of event streams. To address this gap, we introduce \textit{eNavi}, a first-of-its-kind event-based real-world indoor person-following dataset, collected using a TurtleBot $2$ robot, featuring synchronized raw event streams, RGB frames, and expert control actions across indoor maps and trajectories under both normal and low-light conditions. We further build a multi-sensor data processing pipeline that temporally aligns event and RGB observations while reconstructing ground-truth actions from odometry to support high-quality supervised imitation learning. Building on this dataset, we propose a late-fusion RGB-Event navigation policy that combines dual MobileNet encoders with a transformer-based fusion module trained via behavioral cloning. A systematic evaluation of RGB-only, Event-only, and RGB-Event fusion models across $12$ training variations 
ranging from constrained single-path to diverse multi-path settings, shows that policies incorporating event data, particularly the fusion model, achieve improved robustness and lower action prediction error, especially in unseen low-light conditions where RGB-only models fail. We release the dataset, synchronization pipeline, and trained models at \url{https://eventbasedvision.github.io/eNavi/}.

\end{abstract}
\section{Introduction}
\label{sec:intro}

\begin{table*}[]
\centering
\caption{Summary of event-based datasets used in robotic perception and navigation tasks.}
\label{tab:related}
\resizebox{0.72\linewidth}{!}{%
\begin{tabular}{@{}llllcc@{}}
\toprule
\rowcolor[HTML]{FFFFFF} 
{\color[HTML]{1F1F1F} \textbf{Dataset}} &
  {\color[HTML]{1F1F1F} \textbf{Environment}} &
  {\color[HTML]{1F1F1F} \textbf{Sensor Platform}} &
  {\color[HTML]{1F1F1F} \textbf{Primary Task}} &
  {\color[HTML]{1F1F1F} \textbf{Contains Action ($v$, $\omega$)?}} &
  {\color[HTML]{1F1F1F} \textbf{Low-Light Focus?}} \\ \midrule
{\color[HTML]{1F1F1F} MVSEC \cite{zhu2018multivehicle}} &
  {\color[HTML]{1F1F1F} Indoor/Outdoor} &
  {\color[HTML]{1F1F1F} Drone / Car} &
  {\color[HTML]{1F1F1F} SLAM / Pose} &
  {\color[HTML]{1F1F1F} \textbf{×}} &
  {\color[HTML]{1F1F1F} ×} \\
{\color[HTML]{1F1F1F} DDD17 \cite{binas2017ddd17}} &
  {\color[HTML]{1F1F1F} Outdoor} &
  {\color[HTML]{1F1F1F} Car} &
  {\color[HTML]{1F1F1F} Steering Prediction} &
  {\color[HTML]{1F1F1F} \textbf{\checkmark}} &
  {\color[HTML]{1F1F1F} ×} \\
{\color[HTML]{1F1F1F} MMID \cite{bugueno2024mmid}} &
  {\color[HTML]{1F1F1F} Indoor} &
  {\color[HTML]{1F1F1F} Humanoid} &
  {\color[HTML]{1F1F1F} Depth Estimation} &
  {\color[HTML]{1F1F1F} \textbf{×}} &
  {\color[HTML]{1F1F1F} ×} \\
{\color[HTML]{1F1F1F} SEBVS \cite{vinod2025sebvs}} &
  {\color[HTML]{1F1F1F} Synthetic} &
  {\color[HTML]{1F1F1F} Mobile Robot} &
  {\color[HTML]{1F1F1F} Navigation} &
  {\color[HTML]{1F1F1F} \textbf{\checkmark}} &
  {\color[HTML]{1F1F1F} \textbf{×}} \\ \hline
{\color[HTML]{1F1F1F} \textbf{\textit{eNavi} (Ours)}} &
  {\color[HTML]{1F1F1F} \textbf{Indoor}} &
  {\color[HTML]{1F1F1F} \textbf{TurtleBot (Mobile)}} &
  {\color[HTML]{1F1F1F} \textbf{Navigation}} &
  {\color[HTML]{1F1F1F} \textbf{\checkmark}} &
  {\color[HTML]{1F1F1F} \checkmark} \\ \bottomrule
\end{tabular}%
}
\end{table*}

Indoor robot navigation remains a major challenge in service robotics, where robots must operate reliably in environments characterized by clutter, dynamic human motion, and highly variable lighting \cite{shravan2023innovative}. Tasks such as person following demand responsive perception and control under conditions that frequently push conventional sensing to its limits. In particular, frame-based RGB cameras struggle with motion blur, extreme dynamic range scenes, and dim illumination conditions commonly encountered in homes, offices, and public indoor spaces. These limitations highlight the need for sensing pipelines capable of maintaining perceptual quality in visually degraded environments.

Event cameras have recently emerged as a promising sensing modality for robot perception \cite{Gallego2022Event, chakravarthi2024recent, vinod2025sebvs}. Unlike RGB cameras, which capture images at fixed frame rates, event cameras record asynchronous brightness changes with microsecond latency and exceptionally high dynamic range. Their ability to capture fine-grained motion cues under fast movement and poor illumination has enabled substantial progress in low-level perception tasks, including visual odometry \cite{klenk2024deep}, optical flow \cite{shiba2022secrets, liu2022edflow}, and SLAM \cite{tenzin2024application}. However, most existing work focuses on reconstructing scene geometry or motion, while the potential of event-based vision for end-to-end navigation and control remains underexplored.

Despite recent advances, there is no prior dataset that pairs event streams with high-quality action labels in realistic navigation settings. Likewise, research on event-based imitation learning remains limited, particularly for tasks requiring closed-loop robustness under varying illumination.
In this work, we address these gaps by introducing a new dataset and an imitation learning framework tailored to event-based indoor navigation. The dataset captures real-world person-following trajectories across diverse indoor layouts and lighting conditions, providing synchronized RGB frames, raw event streams, and expert control actions. To enable learning from RGB and event modalities, we design a multimodal data preprocessing pipeline that temporally aligns event and RGB observations and reconstructs ground-truth actions from odometry for high-quality imitation learning. Building on this foundation, we present a late-fusion architecture that integrates event-based and RGB cues using lightweight convolutional encoders and a transformer-based cross-modal fusion module.

Beyond the dataset and architecture, we conduct a comprehensive evaluation across multiple sensing configurations under different lighting conditions. By comparing RGB-only, event-only, and fusion-based policies across $12$ training variations, we analyze the contribution of each modality to navigation robustness, particularly in low-light settings where RGB vision fails. These studies reveal the conditions under which event data provides advantages and demonstrate the effectiveness of multimodal fusion for end-to-end control. This work aims to bridge the gap between event-based perception and closed-loop robot navigation. We summarize our contributions below.

\begin{itemize}
    \item \textbf{\textit{A real-world \textit{RGB-Event} indoor navigation dataset}} featuring synchronized RGB frames, raw event streams, and expert control actions collected across multiple maps, trajectories, and illumination conditions (normal and low light).

    \item \textbf{\textit{A multimodal processing pipeline}} that temporally aligns asynchronous event data with RGB observations and reconstructs odometry-based ground-truth actions to enable high-quality supervised imitation learning.

    \item \textbf{\textit{A late-fusion \textit{RGB-Event} navigation policy}} using dual MobileNet encoders and a transformer-based fusion module, along with a comprehensive evaluation of \textit{RGB-only}, \textit{Event-only}, and fusion models across $12$ training regimes demonstrating superior low-light robustness.
\end{itemize}

The remainder of this paper is organized as follows. Section~\ref{sec:related} reviews event-based vision for robot perception and navigation tasks and summarizes relevant policy learning approaches. Section~\ref{sec:method} presents the \textit{eNavi} dataset, including the collection setup, synchronization pipeline, and ground-truth action reconstruction. Section~\ref{sec:enp} introduces the navigation policy, describes the late-fusion architecture, and outlines the training procedure and policy variants. Section~\ref{sec:experiments} reports our experimental setup and results used to validate the proposed hypotheses. Finally, Section~\ref{sec:conclusion} summarizes the key findings and discusses directions for future work.
\begin{figure*}[h!]
  \centering
 
  \includegraphics[width=0.84\linewidth]{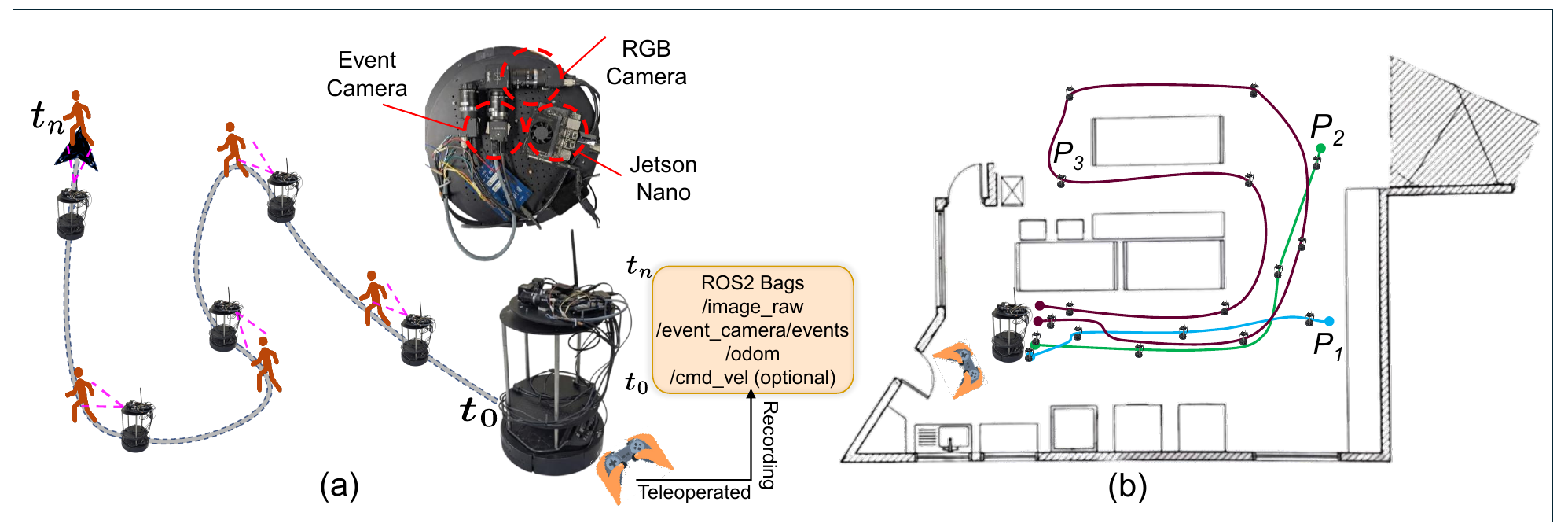}
  \caption{Overview of the dataset generation workflow. \textbf{(a)} The process begins with teleoperated data collection using a mobile robot setup. \textbf{(b)} The collection process where the user is teleoperating the bot to follow another person across the room in three different paths ($P_1$, $P_2$, and $P_3$) in different lighting conditions.}
  \label{fig:fig1}
\end{figure*}

\section{Related Works}
\label{sec:related}

Event-based vision has been extensively investigated across a broad spectrum of application domains, including intelligent transportation systems~\cite{aliminati2024sevd, verma2024etram}, robotic navigation and manipulation, mobility analysis, pose estimation, and dynamic object tracking \cite{chakravarthi2023event, chanda2025event, shravan2023innovative, chen2024syntrac, chanda2025sepose, tan2025real, vinod2026eskitb}.
Building upon these developments, research relevant to our work primarily lies at the intersection of three key areas: event-based robotic perception, multimodal fusion for robust navigation, and end-to-end policy learning for control. In the following section, we review representative contributions in each of these areas and identify the gap that motivates our proposed dataset and policy framework.

\subsection{Event-based Vision and Datasets for Robotic Perception}
\label{sec:related_event}

Recent surveys provide a comprehensive overview of event-based vision methods for perception \cite{chakravarthi2024recent}. Representative works include event-based object detection \cite{wang2024event} and event-based NeRF-style scene representations \cite{hwang2023ev}, demonstrating that asynchronous event streams can support high-quality recognition and 3D reconstruction. 
For robotic applications, event cameras have been particularly effective for perception tasks that benefit from high-speed motion sensing and robustness to extreme illumination, such as visual odometry, optical flow estimation, and SLAM. These methods primarily exploit event data to build low-level geometric or semantic representations of the scene. However, they generally stop at perception and do not directly address learning closed-loop control policies for navigation.

Several event-based datasets have been released over the years~\cite{vinod2026eskitb,vinod2025sebvs,boretti2023pedro}. MVSEC~\cite{zhu2018multivehicle} is a seminal benchmark collected on drones, cars, and handheld platforms, providing event streams, pose, and IMU ground truth for 3D state estimation and SLAM. DSEC~\cite{gehrig2021dsec} and DDD17~\cite{binas2017ddd17} extend this line of work to high-resolution automotive scenarios,  focusing on outdoor driving. These datasets are valuable for perception research, but are recorded in outdoor environments with kinematics and scene structure that differ significantly from indoor mobile robotics.
For datasets such as MMID~\cite{bugueno2024mmid} and PED-Ro~\cite{boretti2023pedro}, which provide synchronized RGB and event streams, are well-suited for tasks like depth estimation or person detection. However, they lack synchronized differential-drive control signals (linear and angular velocity) or wheel odometry aligned with the visual data, making them unsuitable for learning end-to-end navigation or control policies. 
Synthetic alternatives, including EventScape~\cite{gehrig2020video} and SEBVS~\cite{vinod2025sebvs}, utilize simulation frameworks to generate event data alongside ground-truth navigation commands. While these datasets are valuable for controlled experiments, they suffer from sim-to-real gaps, particularly in modeling the noise characteristics and response to illumination behavior of real-world event sensors. A detailed comparison of these datasets with our \textit{eNavi} is provided in Table~\ref{tab:related}.

\subsection{Multimodal Fusion for Robust Robot Navigation}
\label{sec:related_multi}

To overcome the limitations of single-sensor systems, many recent work explores multi-sensor systems for robust robot navigation. For example, fusing RGB with LiDAR has been shown to improve both perception and navigation performance in challenging environments \cite{surmann2020deep}. Multimodal architectures typically combine complementary cues, such as geometry from depth/LiDAR and appearance from RGB, within a unified learning framework.
In contrast, the integration of event cameras into multimodal navigation policies remains relatively rare. Most existing studies that leverage event data do so in perception-only modules or simulation-focused settings, without demonstrating real-world control policies that benefit from event-driven sensing \cite{maqueda2018event}. This leaves open questions about how best to combine event streams with frame-based vision for closed-loop navigation in real environments.

\subsection{Policy Learning for Robot Navigation}
\label{sec:related_policy}

End-to-end policy learning with RGB cameras has emerged as a popular approach for robot navigation. Methods such as \cite{surmann2020deep, bojarski2016end} map raw visual observations directly to control commands using convolutional or transformer-based networks trained via imitation learning or reinforcement learning. These approaches simplify the traditional perception-planning pipeline and have shown strong performance in structured environments.
In the event domain, however, analogous end-to-end navigation policies are still limited. Existing methods \cite{bugueno2025human, vinod2025sebvs} often rely on decoupled perception-and-control pipelines or purely synthetic environments, where the control policy is trained on simulated event data. To the best of our knowledge, there are currently no end-to-end imitation policies trained on real-world \textit{RGB-Event} data for indoor navigation with differential-drive robots. This gap motivates our work, which combines a real-world \textit{RGB-Event} dataset, a multimodal data processing pipeline, and a late-fusion policy architecture to enable event-driven imitation learning for indoor navigation.

% \end{Method}
\section{The \textit{eNavi} Dataset}
\label{sec:method}

\subsection{Hardware and Camera setup}

The data collection platform is built on a \textit{TurtleBot~$2$} mobile robot (Kobuki base), with all cameras and computing units rigidly mounted to the top plate of the chassis, as illustrated in Figure~\ref{fig:fig1}. Onboard processing is handled by an NVIDIA Jetson Orin Nano~\cite{JetsonOrinNano}, running a robot operating system~$2$ (ROS~$2$) for sensor drivers and data logging.
The sensing suite consists of a heterogeneous camera configuration. To minimize parallax, both cameras are mounted on a beam splitter setup as shown Figure~\ref{fig:fig1}(a). The primary sensor is a Prophesee Metavision EVK4 event camera featuring the IMX636 sensor with a resolution of $1280 \times 720$, capable of capturing event data at high temporal resolution under $220\,\mu\text{s}$ latency at $1$k lux. The secondary sensor is a Teledyne Blackfly S USB3 RGB camera equipped with the onsemi Python1300 sensor that captures frames at a $1280\times1024$ resolution up to $170$ frames per second. Co-axial alignment is achieved by mounting both cameras to a THORLabs CCM1-BS013 non-polarizing cube beam splitter. 

\begin{figure}[t]
  \centering
\includegraphics[width=0.95\linewidth]{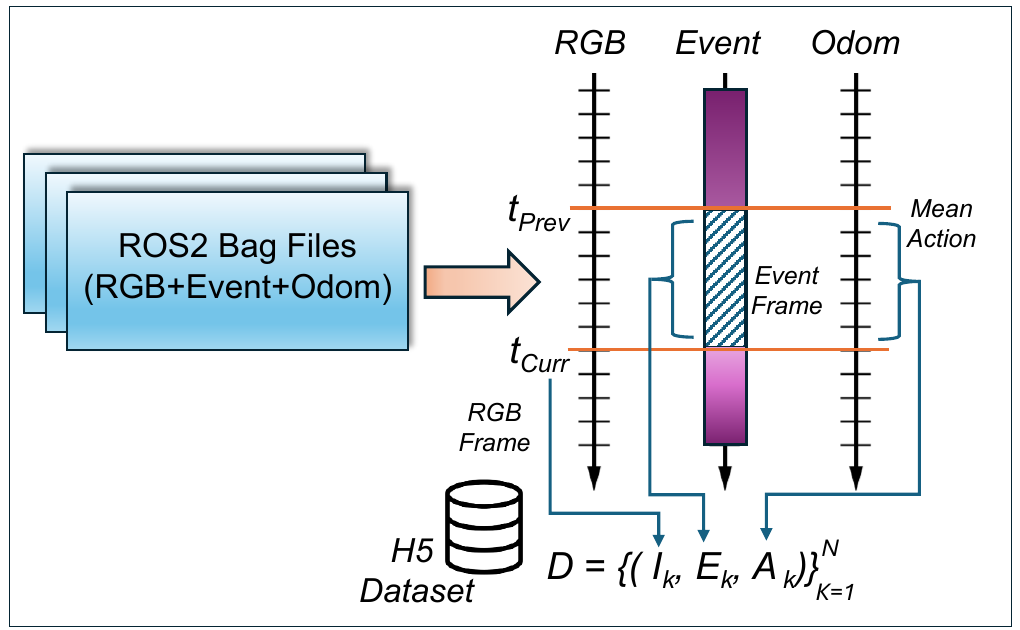}
  \caption{Overview of the dataset generation workflow. The process begins with teleoperated data collection using a mobile robot setup. The recorded ROS $2$ bags are analyzed for trajectory and velocity consistency (center panels) before being synchronized into paired image, event, and action tuples $\{(\mathbf{I}_k, \mathbf{E}_k, \mathbf{a}_k)\}$ stored in .h5 format.}
  \label{fig:fig2}
\end{figure}

Additionally, both cameras were made to share the same field of view using Kowa lenses with $25$mm focal length and maximum aperture of $f/1.4$, compatible with the beam splitter mount cage shown in Figure~\ref{fig:fig1}.
The cameras are time-synchronized at the hardware level by utilizing the onboard external oscillator on an STM32F407 microcontroller to externally trigger both the RGB and event cameras. The microcontroller was configured to send a $30Hz$ pulse with $50\%$ duty cycle to both cameras. This signal is used to both determine the exposure duration of the RGB camera as well as simultaneously inject trigger markers in the event stream, which have been used to temporally align the event and RGB data in post-processing, as depicted in Figure~\ref{fig:fig2}. For spatial correspondence, the cameras were calibrated by using a blinking checkerboard pattern to compute a homography that transforms the RGB frames to the frame of the event camera.
The event camera interfaces with ROS~$2$ via the \texttt{metavision\_driver}, while the RGB camera uses standard \texttt{V4L2} drivers.
For connectivity, a TP-Link Wi-Fi dongle is attached to the Jetson Orin Nano. This wireless link enables remote development access via SSH and facilitates manual teleoperation of the robot using a joystick controller during data collection.

\subsection{Data Collection Protocol}

To ensure generalization, we curate the dataset across three distinct indoor environments. We employ two different human subjects to introduce variability in clothing, body shape, and gait. The subjects are instructed to vary their walking speeds and trajectories to capture a broad range of motion dynamics.
The full dataset comprises approximately two hours of driving data, segmented into over $175$ episodes. To facilitate benchmarking and training, we use single map for in-distribution training and testing, while the remaining two environments are reserved for generalization experiments. This split allows us to evaluate performance both within seen environments and in previously unseen layouts. We plan to release the dataset as an open-source contribution to the research community.

\noindent\textbf{\textit{Collection protocols.}}
To ensure high-quality expert demonstrations, we establish the following protocols during data acquisition:
\begin{itemize}
    \item \textbf{\textit{Privacy and pose:}} To address privacy concerns and maintain a consistent following perspective, the target subject is instructed to keep their back facing the robot at all times. 
    \item \textbf{\textit{Expert teleoperation:}} The robot is teleoperated by a human expert who controls the base using a joystick while viewing the live video feed from the onboard RGB camera. This \textit{first-person-view} (FPV) teleoperation ensures that the collected trajectories represent valid, obstacle-free paths based on visual input.
\end{itemize}

\subsection{Multimodal Data Processing Pipeline}
\label{subsec:processing}

Raw data is recorded in  \texttt{rosbag2} format to enable low-overhead logging of all sensor and control streams. For efficient training and evaluation, these bags are post-processed into structured HDF5 containers that store synchronized image, event, and action tuples. We describe the temporal synchronization methodology and the representation used for event and control data.

\subsubsection{Temporal Synchronization of RGB, Events, and Actions}
\label{subsubsec:sync}

A core challenge in this setup is aligning asynchronous data streams with frame-based observations. The event camera produces a continuous stream of asynchronous events $\mathcal{E}$, while the RGB camera operates at a fixed frame rate, producing frames $\mathcal{I}$, and the robot chassis publishes control commands $\mathcal{C}$ at discrete intervals. We use the RGB frame timestamps as the temporal reference (anchor), as illustrated in Figure~\ref{fig:fig2}.
Let the sequence of recorded RGB frames be denoted as $\mathcal{I} = \{I_1, I_2, \dots, I_N\}$ with corresponding timestamps $T_{\text{rgb}} = \{t_1, t_2, \dots, t_N\}$.

\noindent\textbf{\textit{Event alignment.}}
The continuous event stream is defined as $\mathcal{E} = \{e_k\}_{k=1}^M$, where each event $e_k = (x_k, y_k, t_k, p_k)$ consists of pixel coordinates, timestamp, and polarity. For every RGB frame $I_i$ at time $t_i$, we aggregate the events that occurred in the interval since the previous frame. The set of synchronized events $\mathcal{E}_i$ associated with frame $I_i$ is defined as:
\begin{equation}
    \mathcal{E}_i = \{ e_k \in \mathcal{E} \mid t_{i-1} < t_k \leq t_i \},
\end{equation}
where $t_0$ is the start of the recording. These raw events $\mathcal{E}_i$ are then converted into a dense tensor representation (event frame) to match the spatial dimensions of the RGB input.

\noindent\textbf{\textit{Event representation.}}
To enable feature extraction via a neural network, the raw event set $\mathcal{E}_i$ is transformed into a dense representation $\mathbf{E}_i \in \mathbb{R}^{H \times W \times 2}$ using a spatial event histogram, separating events by polarity (ON/OFF) into two channels.
For a pixel location $(x, y)$, the values for the positive channel ($c=0$) and negative channel ($c=1$) are computed by accumulating events occurring within the synchronization window $\Delta t = t_i - t_{i-1}$:
\begin{equation}
    \mathbf{E}_i(x, y, 0) = \sum_{e_k \in \mathcal{E}_i} \mathbb{I}(x_k = x, y_k = y, p_k = 1),
\end{equation}
\begin{equation}
    \mathbf{E}_i(x, y, 1) = \sum_{e_k \in \mathcal{E}_i} \mathbb{I}(x_k = x, y_k = y, p_k = 0),
\end{equation}
where $\mathbb{I}(\cdot)$ is the indicator function. This yields a $2$-channel input tensor suitable for the event encoder.

\begin{figure*}[t]
  \centering

\includegraphics[width=0.88\linewidth]{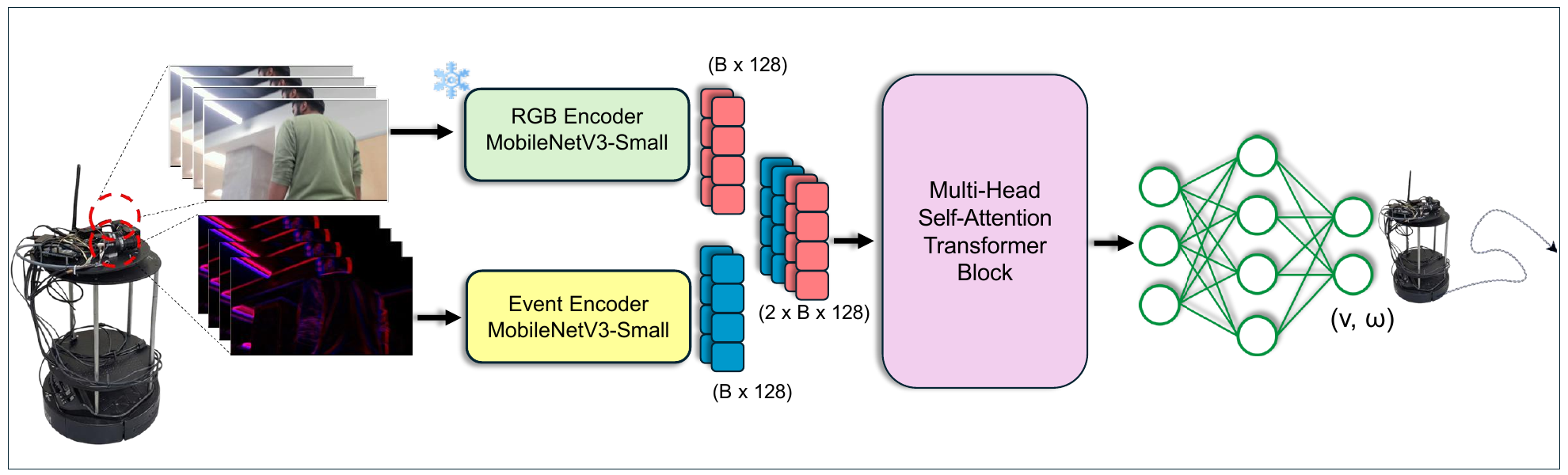}
  \caption{The Event-based Navigation Policy (\textit{ENP-Fusion}) Architecture. The model fuses a frozen RGB stream and a trainable event stream using a Transformer-based attention mechanism. The system takes synchronized \textit{RGB} frames ($320 \times 180 \times 3$ ) and event frames ($320 \times 180 \times 2$) as input and outputs continuous differential drive commands ($v, \omega$) via an MLP head. The snowflake icon denotes that the RGB backbone weights remain fixed during training to prevent overfitting in low-light scenarios.}
  \label{fig:archieture}
\end{figure*}

\noindent\textbf{\textit{Control alignment.}}
Robot control commands (linear velocity $v$ and angular velocity $\omega$) are published asynchronously. To pair a control action with the $i$-th visual observation, we aggregate all commands received between the previous frame timestamp $t_{i-1}$ and the current frame timestamp $t_i$. We compute the mean action $\mathbf{a}_i$ over this interval to obtain a smooth and representative label:
\begin{equation}
    \mathbf{a}_i = \frac{1}{|C_i|} \sum_{c \in C_i} c, 
    \quad \text{where } C_i = \{c_t \mid t_{i-1} < t \leq t_i\}.
\end{equation}
This process yields a synchronized tuple $(\mathbf{I}_i, \mathbf{E}_i, \mathbf{a}_i)$ for every step $i$ in the dataset.

\subsubsection{Odometry-based Action Reconstruction}
\label{subsubsec:command_reconstruction}

A critical requirement for training end-to-end imitation policies is high-fidelity ground truth for the robot's actions. Although the dataset is collected via teleoperation, relying solely on raw joystick input signals can introduce noise due to controller dead zones, communication latency, and operator jitter.
To ensure that the training labels reflect the actual physical motion of the robot, we derive the ground-truth linear ($v$) and angular ($\omega$) velocities directly from the robot's wheel odometry feedback (\texttt{/odom}). Unlike raw control commands, the odometry stream provides an executed estimate of the robot's kinematics.
Rather than differentiating pose changes, we extract instantaneous velocity reports (\texttt{twists}) from the odometry messages. As in the synchronization step, these velocity samples are aggregated over the time window associated with each RGB frame, and their mean is computed to generate the final label $\mathbf{a}_i$. This results in a temporally consistent set of action labels that accurately reflect the robot's executed motion for each synchronized \textit{RGB-event} observation pair.

\label{sec:command}

\section{Event-based Navigation Policy}
\label{sec:enp}

Leveraging the \textit{eNavi} dataset described in Section~\ref{sec:method}, we introduce an event-based navigation policy (\textit{ENP}) for indoor person-following. Unlike traditional unimodal imitation learning approaches, the high temporal resolution and asynchronous nature of event data motivate an architecture that can effectively fuse heterogeneous visual streams. Our framework adopts a hybrid CNN-Transformer design inspired by multimodal sensor-fusion techniques such as TransFuser~\cite{prakash2021multi} and TokenFusion~\cite{wang2022multimodal}, as well as recent work on event-driven servoing policies~\cite{vinod2025sebvs}.

\subsection{The ENP Architecture}
\label{subsec:architecture}

The \textit{ENP} architecture employs a late-fusion strategy to process synchronous RGB and event-frame tensors. As illustrated in Figure~\ref{fig:archieture}, the model consists of two stages: \textbf{(\textit{i})} modality-specific encoding and \textbf{(\textit{ii})} attention-based fusion followed by a control head.

\subsubsection{Modality-specific Encoding}

We use two parallel convolutional encoders to extract feature embeddings from each modality:
\begin{itemize}
    \item \textbf{\textit{RGB encoder.}} A pre-trained MobileNetV3-Small backbone processing the $3$-channel RGB frames to output feature vector. To preserve RGB feature representations of pretrained weights, the weights of the RGB encoder are kept \emph{frozen} during training.
    
    \item \textbf{\textit{Event encoder.}} A MobileNetV3-Small backbone which processes the $2$-channel event representation described in Section~\ref{subsec:processing}. Since this encoder's weights were not originally trained for events, this encoder is kept trainable, allowing it to learn features tailored to the sparse, high-temporal-resolution event data.
\end{itemize}

\noindent Each encoder outputs a compact feature vector (or token), which is subsequently passed to the fusion module.

\subsubsection{Attention-based Fusion and Control Head}

The feature vectors from the RGB and event encoders are tokenized and concatenated into a unified sequence ($2$ feature tokens) processed by a Transformer encoder block. This self-attention-based fusion mechanism enables the policy to adaptively weigh the contribution of RGB context versus event dynamics on a per-sample basis, which is particularly important under varying illumination and motion conditions.

\noindent The fused representation is fed into a multilayer perceptron (MLP) policy head that predicts continuous differential-drive commands, namely the linear velocity `$v$' and angular velocity `$\omega$'. At inference time, the policy maps each synchronized observation $(\mathbf{I}_i, \mathbf{E}_i)$ to an action $\hat{\mathbf{a}}_i = ( \hat{v}_i, \hat{\omega}_i )$.

\subsection{Policy Variants and Training Methods}
\label{subsec:variants}

To systematically study the contribution of event data under different levels of task complexity and illumination, we train a family of \textit{ENP} variants. The design spans three architectures and four different subsets from the \textit{eNavi} dataset, yielding $12$ model variants.

\subsubsection{The Architecture Variants}

We evaluate three modality configurations:

\begin{itemize}
    \item \textbf{\textit{ENP-RGB}:} A baseline model that uses only the frozen RGB encoder. The event branch is removed, and the policy operates purely on frame-based observations.
    \item \textbf{\textit{ENP-Event}:} A model that relies exclusively on the trainable event encoder, operating purely on the event representation.
    \item \textbf{\textit{ENP-Fusion}:} The proposed multimodal model that uses both encoders and the Transformer fusion block to attend jointly to RGB and event features.
\end{itemize}

These variants allow us to isolate the benefits of event-only sensing and \textit{RGB-Event} fusion relative to a conventional \textit{RGB-only} baseline.

\begin{table*}[t]
\centering
\caption{Validation metrics across all the training.
 All variations are trained for $50$ epochs, where Train* indicates the model was run with early stopping at $8$ epochs. }
\label{tab:val_table}
\resizebox{0.80\textwidth}{!}{%
\begin{tabular}{c|cccc|c|cccc}
\toprule
\rowcolor[HTML]{FFFFFF} 
\textbf{Variation} &
  \textbf{Modal} &
  \textbf{Train*} &
  \textbf{Best} &
  \textbf{MAE} &
  \textbf{Variation} &
  \textbf{Modal} &
  \textbf{Train*} &
  \textbf{Best} &
  \textbf{MAE} \\ \midrule
\rowcolor[HTML]{FFFFFF} 
\cellcolor[HTML]{FFFFFF} & RGB   & 50 & 48 & 0.0548 & \cellcolor[HTML]{FFFFFF} & RGB   & 50 & 50 & 0.0667 \\
\rowcolor[HTML]{FFFFFF} 
\cellcolor[HTML]{FFFFFF} & Event & 37 & 29 & 0.0393 & \cellcolor[HTML]{FFFFFF} & Event & 44 & 36 & \textbf{0.0378} \\
\rowcolor[HTML]{FFFFFF} 
\multirow{-3}{*}{\cellcolor[HTML]{FFFFFF}\textbf{ (Single-Path, Normal-Light)}} &
  Fusion &
  27 &
  19 &
  \textbf{0.0388} &
  \multirow{-3}{*}{\cellcolor[HTML]{FFFFFF}\textbf{(Multi-Path, Normal-Light)}} &
  Fusion &
  29 &
  21 &
  0.0402 \\ \hline
\rowcolor[HTML]{FFFFFF} 
\cellcolor[HTML]{FFFFFF} & RGB   & 50 & 50 & 0.0584 & \cellcolor[HTML]{FFFFFF} & RGB   & 50 & 50 & 0.0707 \\
\rowcolor[HTML]{FFFFFF} 
\cellcolor[HTML]{FFFFFF} & Event & 26 & 18 & 0.0388 & \cellcolor[HTML]{FFFFFF} & Event & 18 & 10 & 0.0416 \\
\rowcolor[HTML]{FFFFFF} 
\multirow{-3}{*}{\cellcolor[HTML]{FFFFFF}\textbf{(Single-Path, Mixed-Light)}} &
  Fusion &
  31 &
  23 &
  \textbf{0.0358} &
  \multirow{-3}{*}{\cellcolor[HTML]{FFFFFF}\textbf{(Multi-Path, Mixed-Light)}} &
  Fusion &
  30 &
  22 &
  \textbf{0.0370} \\ \bottomrule
\end{tabular}%
}
\end{table*}

\begin{table*}[t]
\centering
\caption{Evaluation of \textit{Normal-Light} under varying illumination. The \textit{YOLOv8n + PID} baseline is evaluated on its respective path type \textit{Single-Path} (top) and \textit{Multi-Path} (down) for direct comparison.}
\label{tab:test_table}
\resizebox{0.80\textwidth}{!}{%
\begin{tabular}{c|c|ccc|ccc}
\toprule
\rowcolor[HTML]{FFFFFF} 
\cellcolor[HTML]{FFFFFF} &
  \cellcolor[HTML]{FFFFFF} &
  \multicolumn{3}{c|}{\cellcolor[HTML]{FFFFFF}\textbf{Normal-Light}} &
  \multicolumn{3}{c}{\cellcolor[HTML]{FFFFFF}\textbf{Low-Light}} \\ \cline{3-8} 
\rowcolor[HTML]{FFFFFF} 
\multirow{-2}{*}{\cellcolor[HTML]{FFFFFF}\textbf{Method}} &
  \multirow{-2}{*}{\cellcolor[HTML]{FFFFFF}\textbf{Modal}} &
  \multicolumn{1}{c|}{\cellcolor[HTML]{FFFFFF}\textbf{Linear MAE}} &
  \multicolumn{1}{c|}{\cellcolor[HTML]{FFFFFF}\textbf{Angular MAE}} &
  \textbf{Total MAE} &
  \multicolumn{1}{c|}{\cellcolor[HTML]{FFFFFF}\textbf{Linear MAE}} &
  \multicolumn{1}{c|}{\cellcolor[HTML]{FFFFFF}\textbf{Angular MAE}} &
  \textbf{Total MAE} \\ \midrule
\rowcolor[HTML]{FFFFFF} 
\textbf{Yolo8n + PID  (Single-Path)} &
  RGB &
  \multicolumn{1}{c|}{\cellcolor[HTML]{FFFFFF}0.0882} &
  \multicolumn{1}{c|}{\cellcolor[HTML]{FFFFFF}0.0756} &
  0.0819 &
  \multicolumn{1}{c|}{\cellcolor[HTML]{FFFFFF}0.0758} &
  \multicolumn{1}{c|}{\cellcolor[HTML]{FFFFFF}0.0667} &
  0.0713 \\ \hline
\rowcolor[HTML]{FFFFFF} 
\cellcolor[HTML]{FFFFFF} &
  \cellcolor[HTML]{FFFFFF}RGB &
  \multicolumn{1}{c|}{\cellcolor[HTML]{FFFFFF}0.0182} &
  \multicolumn{1}{c|}{\cellcolor[HTML]{FFFFFF}0.0237} &
  \textbf{0.0210} &
  \multicolumn{1}{c|}{\cellcolor[HTML]{FFFFFF}0.0258} &
  \multicolumn{1}{c|}{\cellcolor[HTML]{FFFFFF}0.0168} &
  \textbf{0.0213} \\
\rowcolor[HTML]{FFFFFF} 
\cellcolor[HTML]{FFFFFF} &
  Event &
  \multicolumn{1}{c|}{\cellcolor[HTML]{FFFFFF}0.0177} &
  \multicolumn{1}{c|}{\cellcolor[HTML]{FFFFFF}0.0275} &
  0.0266 &
  \multicolumn{1}{c|}{\cellcolor[HTML]{FFFFFF}0.0234} &
  \multicolumn{1}{c|}{\cellcolor[HTML]{FFFFFF}0.0236} &
  0.0235 \\
\rowcolor[HTML]{FFFFFF} 
\multirow{-3}{*}{\cellcolor[HTML]{FFFFFF}\textbf{ Single-Path, Normal-Light (Ours)}} &
  Fusion &
  \multicolumn{1}{c|}{\cellcolor[HTML]{FFFFFF}0.0183} &
  \multicolumn{1}{c|}{\cellcolor[HTML]{FFFFFF}0.0259} &
  0.0221 &
  \multicolumn{1}{c|}{\cellcolor[HTML]{FFFFFF}0.0246} &
  \multicolumn{1}{c|}{\cellcolor[HTML]{FFFFFF}0.0199} &
  0.0222 \\ \hline
\rowcolor[HTML]{FFFFFF} 
\textbf{Yolo8n + PID (Multi-Path)} &
  RGB &
  \multicolumn{1}{c|}{\cellcolor[HTML]{FFFFFF}0.0841} &
  \multicolumn{1}{c|}{\cellcolor[HTML]{FFFFFF}0.1155} &
  0.0998 &
  \multicolumn{1}{c|}{\cellcolor[HTML]{FFFFFF}0.0724} &
  \multicolumn{1}{c|}{\cellcolor[HTML]{FFFFFF}0.1342} &
  0.1033 \\ \hline
\rowcolor[HTML]{FFFFFF} 
\cellcolor[HTML]{FFFFFF} &
  RGB &
  \multicolumn{1}{c|}{\cellcolor[HTML]{FFFFFF}0.0221} &
  \multicolumn{1}{c|}{\cellcolor[HTML]{FFFFFF}0.0805} &
  0.0463 &
  \multicolumn{1}{c|}{\cellcolor[HTML]{FFFFFF}0.0128} &
  \multicolumn{1}{c|}{\cellcolor[HTML]{FFFFFF}0.0901} &
  0.0514 \\
\rowcolor[HTML]{FFFFFF} 
\cellcolor[HTML]{FFFFFF} &
  Event &
  \multicolumn{1}{c|}{\cellcolor[HTML]{FFFFFF}0.0090} &
  \multicolumn{1}{c|}{\cellcolor[HTML]{FFFFFF}0.0519} &
  \cellcolor[HTML]{FFFFFF}\textbf{0.0305} &
  \multicolumn{1}{c|}{\cellcolor[HTML]{FFFFFF}0.0157} &
  \multicolumn{1}{c|}{\cellcolor[HTML]{FFFFFF}0.0912} &
  0.0534 \\
\rowcolor[HTML]{FFFFFF} 
\multirow{-3}{*}{\cellcolor[HTML]{FFFFFF}\textbf{Multi-Path, Normal-Light (Ours)}} &
  Fusion &
  \multicolumn{1}{c|}{\cellcolor[HTML]{FFFFFF}0.0091} &
  \multicolumn{1}{c|}{\cellcolor[HTML]{FFFFFF}0.0578} &
  0.0335 &
  \multicolumn{1}{c|}{\cellcolor[HTML]{FFFFFF}0.0155} &
  \multicolumn{1}{c|}{\cellcolor[HTML]{FFFFFF}0.0779} &
  \cellcolor[HTML]{FFFFFF}\textbf{0.0467} \\ \bottomrule
\end{tabular}%
}
\end{table*}

\subsubsection{The Dataset Variants}

To probe robustness to both trajectory diversity and illumination changes, we factor our training data along two axes: \textbf{(\textit{i})} path complexity and \textbf{(\textit{ii})} lighting conditions. Combined with the three architecture variants (Section~\ref{subsec:variants}), this yields $3 \times 2 \times 2 = 12$ trained policies.

\paragraph{\textit{Path complexity}}
We consider two trajectory variants:
\begin{itemize}
    \item \textbf{\textit{Single-Path}:} A single, relatively simple person-following trajectory ($P_1$), as in Figure \ref{fig:fig1}(b), used to study imitation performance in a controlled layout.
    \item \textbf{\textit{Multi-Path}:} A union of three trajectories ($P_1$,$P_2$,$P_3$) as in Figure \ref{fig:fig1}(b), used to evaluate generalization across diverse paths and behaviors.
\end{itemize}
\paragraph{\textit{Lighting conditions.}}
For each path variant, we vary the illumination:
\begin{itemize}
    \item \textbf{\textit{Normal-Light}:} Only trajectories recorded under standard indoor lighting.
    \item \textbf{\textit{Mixed-Light}:} Trajectories recorded under both normal and low-light conditions, representing a more challenging domain with degraded RGB observations.
\end{itemize}

\noindent This factorial design allows us to systematically analyze how architecture choice (\textit{ENP-RGB}, \textit{ENP-Event}, \textit{ENP-Fusion}), path complexity (\textit{Single-Path} vs \textit{Multi-Path}), and lighting (\textit{Normal-Light} vs \textit{Mixed-Light}) interact to affect navigation performance and robustness.

\subsection{Behavioral Cloning Objective}
\label{subsubsec:bc}

We train all \textit{ENP} variants using behavioral cloning (BC), a standard imitation learning method. The problem is formulated as supervised regression, where the goal is to find network parameters `$\theta$' that minimize the discrepancy between the predicted actions and the expert commands.
Given a dataset of $N$ synchronized observation-action pairs
$\mathcal{D} = \{(o_i, u^*_i)\}_{i=1}^{N}$, where $o_i$ denotes the multimodal sensor input (RGB and/or event) and $u^*_i$ represents the expert's linear and angular velocities, we optimize the policy using the mean absolute error (MAE) loss:
\begin{equation}
    \mathcal{L}_{\text{BC}}(\theta) = \frac{1}{N} \sum_{i=1}^{N} \big\| \pi_{\theta}(o_i) - u^*_i \big\|_1,
\end{equation}
where $\pi_{\theta}(o_i)$ is the action predicted by the network for the $i$-th observation. We use the L1 norm (MAE) instead of mean squared error (MSE) to improve robustness to outliers and the high-frequency noise often present in expert command signals.

\label{sec:experiments}

\section{Experiments and Results}
\label{sec:experiments}

In this section, we evaluate the proposed \textit{ENP} on the \textit{eNavi} dataset. Our experimental design is structured around two hypotheses:

\begin{itemize}
    \item \textbf{\textit{Multimodal training efficiency:}} Fusing synchronous RGB context with asynchronous event dynamics leads to faster convergence and lower validation error compared to RGB baselines, which may struggle to extract robust features under challenging conditions.
    
    \item \textbf{\textit{Low-light robustness/Zero-shot generalization:}} Event-driven policies trained exclusively on \textit{Normal-Light} data generalize better to unseen low-light conditions than \textit{RGB-only} baselines, which suffer from texture degradation in the dark.

\end{itemize}
%%%%%%%%%%%%%%%%%%%%%%%%%%%%%%%%%%%%%%%%%%%%%%%

\begin{figure*}[t]
  \centering
\includegraphics[width=0.825\linewidth]{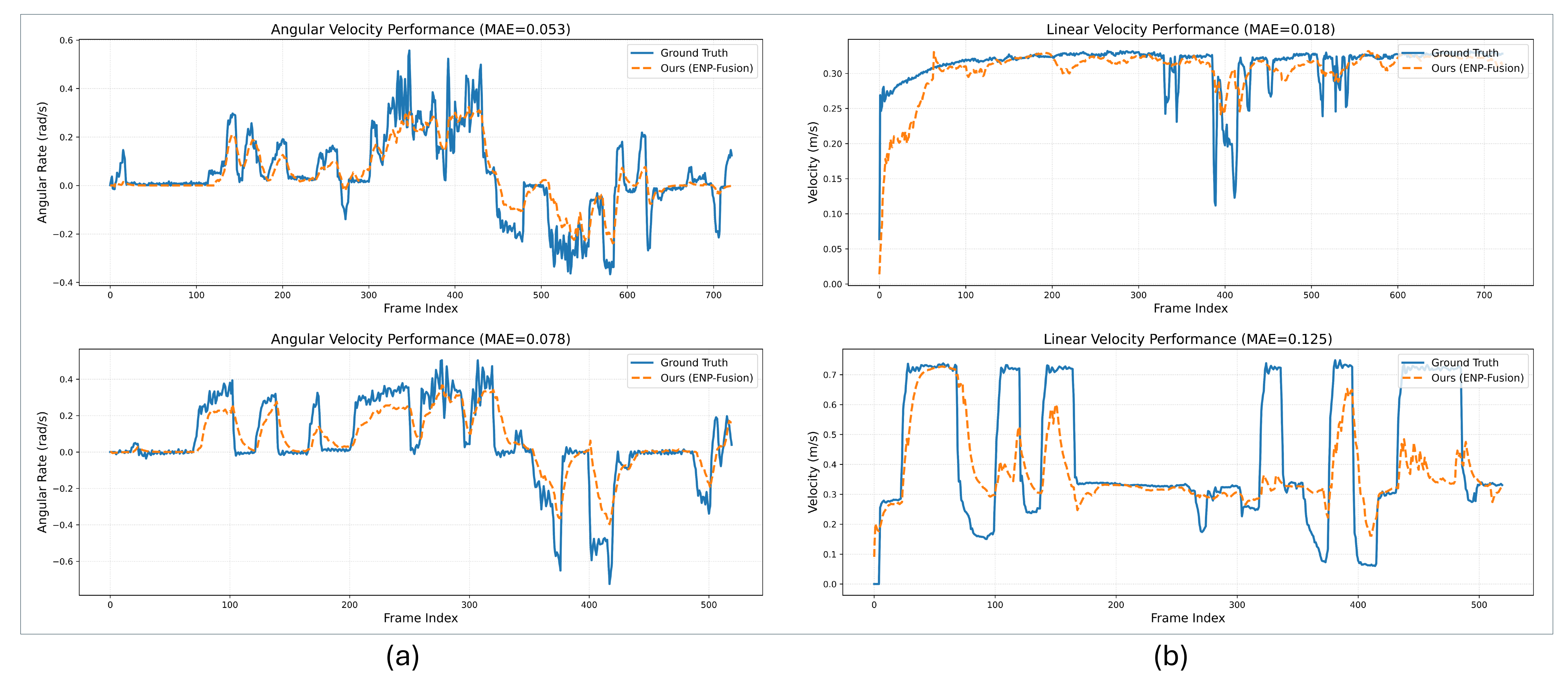}
  \caption{Trajectory Analysis of \textit{ENP-Fusion} Policy. Comparison of predicted linear velocity (m/s) and angular rate (rad/s) against expert ground truth for representative test trajectory ($P_3$). (a) and (b) contrast performance under normal-light (top) and low-light (down) conditions.}
  \label{fig:fig4}
\end{figure*}

\subsection{Training Setup and Convergence}
\label{subsec:setup}

All policy variants are implemented in PyTorch and trained using the AdamW optimizer with a learning rate of $2 \times 10^{-4}$ and a weight decay of $3 \times 10^{-4}$ for regularization. We use a batch size of $64$ and an $80$/$10$/$10$ split for training, validation, and test sets, ensuring that test trajectories are completely held out during training.
To prevent overfitting and to assess learning efficiency across modalities, we employ early stopping with a patience of $8$ epochs and a maximum training budget of $50$ epochs. Validation performance for all $12$ training configurations is summarized in Table~\ref{tab:val_table}. Several trends emerge:

\begin{itemize}
    \item \textbf{\textit{RGB saturation}:} \textit{RGB-only} models consistently run to the full $50$-epoch budget, especially on the more complex (\textit{Multi-Path}, \textit{Mixed-Light}) settings. This suggests that the network has difficulty extracting sufficiently robust control features from RGB inputs alone within the allotted training horizon.
    
    \item \textbf{\textit{Event-driven efficiency}:} Models that incorporate event information (Event-only and Fusion) converge substantially faster, often triggering early stopping between epochs $20$ and $35$. This indicates that the high-temporal-resolution event cues provide a stronger and more informative learning signal.
    
    \item \textbf{\textit{Validation performance gap}:} Fusion and Event-only models achieve the lowest validation MAE in most regimes. For example, in the (Multi-Path, Mixed-Light) setting in Table~\ref{tab:val_table}, \textit{ENP-Fusion} reaches an MAE of $0.0370$, compared to \textit{ENP-RGB}'s $0.0707$. This supports the \textit{multimodal training efficiency} hypothesis, showing that multimodal policy \textit{ENP-Fusion} has faster convergence and better final validation performance compared to other variants.
\end{itemize}

\subsection{Zero-Shot Generalization to Low Light}
\label{subsec:generalization}

To evaluate the \textit{Low-Light Robustness} hypothesis, we consider policies trained \emph{only} on (\textit{Normal-Light}) data and test them on both (\textit{Normal-Light}) and (\textit{Low-Light}) conditions. Table~\ref{tab:test_table} reports results for two scenarios: a simple single-path setting (\textit{Single-Path}, \textit{Normal-Light}) and a more complex multi-path setting (\textit{Multi-Path}, \textit{Normal-Light}), alongside a YOLO8n + PID baseline. Since there are no prior end-to-end event-based navigation policies for real indoor robots, we compare against: (a) a classical perception+control pipeline (YOLOv8n+PID), (b) RGB-only (ENP-RGB) and (c) event-only (ENP-Event), where we trained (b) and (c). These represent the closest comparators in the absence of comparable event-based baselines.
The results reveal a clear distinction between simple and complex navigation:

\begin{itemize}
    \item \textbf{\textit{Simple trajectories (\textit{Single-Path}, \textit{Normal-Light})}:} On the single-path dataset, the \textit{ENP-RGB} policy generalizes well to low light, with Total MAE changing only slightly ($0.0210\rightarrow0.0213$). Fusion and \textit{Event-only} variants perform similarly. This suggests that for simple, structured trajectories, RGB sensors retain sufficient contrast even under reduced illumination.
    
    \item \textbf{\textit{Complex trajectories (Multi-Path, Normal-Light):}} In the \textit{Multi-Path} training, the benefits of fusion become pronounced. The difference from normal to low light, the \textit{ENP-RGB} policy changes from ($0.0463 \rightarrow 0.0514$), while the \textit{ENP-Fusion} policy generalizes well and achieves a lower change ($0.0335 \rightarrow 0.0467$). 
\end{itemize}

\noindent These findings support \textit{Low-light robustness} of fusion policies and \textit{Zero-shot generalization} capabilities of event data.

% \begin{figure}[t]
%   \centering
% \includegraphics[width=0.9\linewidth]{fig/figure05.pdf}
%   \caption{Overlay visualization of Event and RGB data from the Low-Light samples in our dataset, where Red and Blue points represent the event frame, top of the degraded RGB frame in low-light conditions.}
%   \label{fig:fig5}
% \end{figure}

\subsection{Trajectory Analysis}
\label{subsec:results}

To better understand the control characteristics of the best-performing policy, we analyze the \textit{ENP-Fusion}, which has been trained on the (\textit{Multi-Path}, \textit{Mixed-Light}) data. On the test case sample, we plot the difference in Ground Truth from our dataset and our \textit{ENP-Fusion} model.    
Figure~\ref{fig:fig4} visualizes the predicted linear and angular velocities against the expert commands for representative trajectories under normal-light (left) and low-light (right) conditions.

\medskip
\noindent\textbf{Robustness in low light.}
Despite degraded RGB features in the low-light domain,
%as shown in samples in Fig~\ref{fig:fig5},
The \textit{ENP-Fusion} policy maintains stable and accurate control:

\begin{itemize}
    \item \textbf{Linear velocity:} The model successfully tracks high-speed segments ($\approx 0.33$\,m/s) with a low linear MAE of $0.125$ as shown in Figure~\ref{fig:fig4}(b), indicating that forward motion planning remains accurate even under illumination changes.
    
    \item \textbf{Angular velocity:} Similar to linear velocity, steering predictions continue to align with the expert's turns, with an angular MAE of $0.078$. While slightly noisier than in normal-light conditions, the policy captures all major directional changes, confirming that the event stream provides sufficient information for complex maneuvering.
\end{itemize}

\noindent Overall, these quantitative and trajectory-level results validate both our claims. Integrating event-based sensing into end-to-end navigation policies like \textit{ENP-Fusion} leads to faster training, improved generalization, and enhanced robustness in low-light indoor environments compared to \textit{RGB-only} baselines.

\section{Discussion}
\label{sec:discussion}

As demonstrated in Section \ref{sec:experiments}, the \textit{ENP-Fusion} architecture exhibits faster convergence, enhanced robustness, and superior generalization in low-light conditions. We attribute these improvements to the effective integration of high-temporal event data with context-rich RGB frames, which minimizes MAE across all variations.
However, our current data processing pipeline, described in Section~\ref{sec:method}, aligns RGB and event streams to a fixed frequency. Because the policies are trained on this synchronized data, optimizing the current processing pipeline in a physical robot for real-time inference presents significant computational challenges. This challenge limits the practicality of time-synchronized fusion approaches, as combining inference with processing is computationally demanding for edge devices, making single-modality approaches such as \textit{ENP-RGB} and \textit{ENP-Event} more viable for real-time deployment. 

As one of the first works exploring this space, we highlight this limitation and outline key directions for future research. We release our first-of-its-kind \textit{event-rgb-action} dataset \textit{eNavi}, in both raw ROS2 bag and processed H5 formats, precisely to enable the community to develop improved fusion techniques between event and RGB data. As our work demonstrates the utility of the event data for low-light generalization, we believe that improved fusion techniques will bring mobile robots a step closer to reliable low-light navigation. Apart from fusion techniques, further research into asynchronous data processing, like passing event and RGB data at their native, separate intervals, could yield more 
responsive control policies. Additionally, developing policy heads that predict action chunks, for instance, via diffusion-based policy heads, and integrating online learning techniques such as DAgger~\cite{ross2011reduction} could yield smoother and more robust robot trajectories. Ultimately, our dataset and pipeline serve as a stepping-stone for the development of specialized event-based datasets for imitation learning and vision-language models that fully exploit the high temporal resolution of event-based vision.

\section{Conclusion}
\label{sec:conclusion}

This work addresses the underexplored domain of event-based imitation learning policies for robotic control. We investigate novel fusion techniques for integrating event and RGB data and present a new methodology for developing state-action pair datasets incorporating event-based modalities. A primary contribution of this research is the \textit{eNavi} dataset, a first-of-its-kind resource comprising $175$ episodes totaling two hours of data collected from a teleoperated physical robot in diverse indoor environments under varying illumination conditions. Our findings evaluate the specific implications of training policies with multimodal sensor inputs, demonstrating how the integration of event and RGB cameras enhances robustness. The experimental results provide strong evidence that fusion-based policies achieve superior generalization across challenging lighting scenarios. Ultimately, this work establishes a comprehensive pipeline for fusing event and RGB data and introduces effective training frameworks for end-to-end control policies capable of performing human-following tasks.

\bibliographystyle{IEEEtran}
\bibliography{root}

@String(CVPR= {IEEE Conf. Comput. Vis. Pattern Recog.})

@String(CVPR  = {CVPR})

@inproceedings{klenk2024deep,
  title={Deep event visual odometry},
  author={Klenk, Simon and Motzet, Marvin and Koestler, Lukas and Cremers, Daniel},
  booktitle={2024 International conference on 3D vision (3DV)},
  pages={739--749},
  year={2024},
  organization={IEEE}
}

@inproceedings{boretti2023pedro,
  title={Pedro: an event-based dataset for person detection in robotics},
  author={Boretti, Chiara and Bich, Philippe and Pareschi, Fabio and Prono, Luciano and Rovatti, Riccardo and Setti, Gianluca},
  booktitle={Proceedings of the IEEE/CVF Conference on Computer Vision and Pattern Recognition},
  pages={4065--4070},
  year={2023}
}

@inproceedings{wang2024event,
  title={Event stream-based visual object tracking: A high-resolution benchmark dataset and a novel baseline},
  author={Wang, Xiao and Wang, Shiao and Tang, Chuanming and Zhu, Lin and Jiang, Bo and Tian, Yonghong and Tang, Jin},
  booktitle={Proceedings of the IEEE/CVF Conference on Computer Vision and Pattern Recognition},
  pages={19248--19257},
  year={2024}
}

@inproceedings{hwang2023ev,
  title={Ev-nerf: Event based neural radiance field},
  author={Hwang, Inwoo and Kim, Junho and Kim, Young Min},
  booktitle={Proceedings of the IEEE/CVF Winter Conference on Applications of Computer Vision},
  pages={837--847},
  year={2023}
}

@article{bojarski2016end,
  title={End to end learning for self-driving cars},
  author={Bojarski, Mariusz and Del Testa, Davide and Dworakowski, Daniel and Firner, Bernhard and Flepp, Beat and Goyal, Prasoon and Jackel, Lawrence D and Monfort, Mathew and Muller, Urs and Zhang, Jiakai and others},
  journal={arXiv preprint arXiv:1604.07316},
  year={2016}
}

@inproceedings{maqueda2018event,
  title={Event-based vision meets deep learning on steering prediction for self-driving cars},
  author={Maqueda, Ana I and Loquercio, Antonio and Gallego, Guillermo and Garc{\'\i}a, Narciso and Scaramuzza, Davide},
  booktitle={Proceedings of the IEEE Conference on Computer Vision and Pattern Recognition (CVPR)},
  pages={5419--5427},
  year={2018}
}

@inproceedings{prakash2021multi,
  title={Multi-modal fusion transformer for end-to-end autonomous driving},
  author={Prakash, Aditya and Chitta, Kashyap and Geiger, Andreas},
  booktitle={Proceedings of the IEEE/CVF Conference on Computer Vision and Pattern Recognition (CVPR)},
  pages={7077--7087},
  year={2021}
}

@inproceedings{verma2024etram,
  title={etram: Event-based traffic monitoring dataset},
  author={Verma, Aayush Atul and Chakravarthi, Bharatesh and Vaghela, Arpitsinh and Wei, Hua and Yang, Yezhou},
  booktitle={Proceedings of the IEEE/CVF conference on computer vision and pattern recognition},
  pages={22637--22646},
  year={2024}
}

@article{aliminati2024sevd,
  title={Sevd: Synthetic event-based vision dataset for ego and fixed traffic perception},
  author={Aliminati, Manideep Reddy and Chakravarthi, Bharatesh and Verma, Aayush Atul and Vaghela, Arpitsinh and Wei, Hua and Zhou, Xuesong and Yang, Yezhou},
  journal={arXiv preprint arXiv:2404.10540},
  year={2024}
}

@article{tenzin2024application,
  title={Application of event cameras and neuromorphic computing to VSLAM: A survey},
  author={Tenzin, Sangay and Rassau, Alexander and Chai, Douglas},
  journal={Biomimetics},
  volume={9},
  number={7},
  pages={444},
  year={2024}
}

@inproceedings{wang2022multimodal,
  title={Multimodal token fusion for vision transformers},
  author={Wang, Yikai and Huang, Wenbing and Sun, Fuchun and Xu, Tingyang and Rong, Yu and Huang, Junzhou},
  booktitle={Proceedings of the IEEE/CVF Conference on Computer Vision and Pattern Recognition (CVPR)},
  pages={12186--12195},
  year={2022}
}

@article{vinod2025sebvs,
  title={SEBVS: Synthetic Event-based Visual Servoing for Robot Navigation and Manipulation},
  author={Vinod, Krishna and Ramesh, Prithvi Jai and Chakravarthi, Bharatesh and others},
  journal={arXiv preprint arXiv:2508.17643},
  year={2025}
}

@inproceedings{chakravarthi2023event,
  title={Event-based sensing for improved traffic detection and tracking in intelligent transport systems toward sustainable mobility},
  author={Chakravarthi, Bharatesh and Manoj Kumar, M and Pavan Kumar, BN},
  booktitle={International Conference on Interdisciplinary Approaches in Civil Engineering for Sustainable Development},
  pages={83--95},
  year={2023},
  organization={Springer}
}

@inproceedings{chen2024syntrac,
  title={SynTrac: a synthetic dataset for traffic signal control from traffic monitoring cameras},
  author={Chen, Tiejin and Shirke, Prithvi and Chakravarthi, Bharatesh and Vaghela, Arpitsinh and Da, Longchao and Lu, Duo and Yang, Yezhou and Wei, Hua},
  booktitle={2024 IEEE 27th International Conference on Intelligent Transportation Systems (ITSC)},
  pages={2386--2391},
  year={2024},
  organization={IEEE}
}

@article{chanda2025sepose,
  title={Sepose: A synthetic event-based human pose estimation dataset for pedestrian monitoring},
  author={Chanda, Kaustav and Verma, Aayush Atul and Vaghela, Arpitsinh and Yang, Yezhou and Chakravarthi, Bharatesh},
  journal={arXiv preprint arXiv:2507.11910},
  year={2025}
}

@article{zhu2018multivehicle,
  title={The Multi Vehicle Stereo Event Camera Dataset: An Event Camera Dataset for 3D Perception},
  author={Zhu, Alex Zihao and Thakur, Dinesh and Ozaslan, Tolga and Pfrommer, Bernd and Kumar, Vijay and Daniilidis, Kostas},
  journal={IEEE Robotics and Automation Letters (RA-L)},
  volume={3},
  number={3},
  pages={2032--2039},
  year={2018},
  publisher={IEEE}
}

@article{tan2025real,
  title={How real is carlas dynamic vision sensor? a study on the sim-to-real gap in traffic object detection},
  author={Tan, Kaiyuan and Chakravarthi, Bharatesh and others},
  journal={arXiv preprint arXiv:2506.13722},
  year={2025}
}

@article{surmann2020deep,
  title={Deep reinforcement learning for real autonomous mobile robot navigation in indoor environments},
  author={Surmann, Hartmut and Jestel, Christian and Marchel, Robin and Musberg, Franz and Elhadj, Huma and Goulermas, John Y},
  journal={arXiv preprint arXiv:2005.13857},
  year={2020}
}

@inproceedings{chanda2025event,
  title={Event quality score (eqs): Assessing the realism of simulated event camera streams via distance in latent space},
  author={Chanda, Kaustav and Verma, Aayush and Vaghela, Arpitsinh and Yang, Yezhou and Chakravarthi, Bharatesh},
  booktitle={Proceedings of the Computer Vision and Pattern Recognition Conference},
  pages={5105--5113},
  year={2025}
}

@article{vinod2026eskitb,
  title={eSkiTB: A Synthetic Event-based Dataset for Tracking Skiers},
  author={Vinod, Krishna and Vishal, Joseph Raj and Chanda, Kaustav and Ramesh, Prithvi Jai and Yang, Yezhou and Chakravarthi, Bharatesh},
  journal={arXiv preprint arXiv:2601.06647},
  year={2026}
}

@article{gehrig2021dsec,
  title={DSEC: A Stereo Event Camera Dataset for Driving Scenarios},
  author={Gehrig, Mathias and Aarents, Willem and Gehrig, Daniel and Scaramuzza, Davide},
  journal={IEEE Robotics and Automation Letters (RA-L)},
  volume={6},
  number={3},
  pages={4947--4954},
  year={2021},
  publisher={IEEE}
}

@inproceedings{binas2017ddd17,
  title={DDD17: End-To-End DAVIS Driving Dataset},
  author={Binas, Jonathan and Neil, Daniel and Liu, Shih-Chii and Delbruck, Tobi},
  booktitle={ICML Workshop on Machine Learning for Autonomous Vehicles},
  year={2017}
}

@inproceedings{gehrig2020video,
  title={Video to Events: Recycling Video Datasets for Event Cameras},
  author={Gehrig, Daniel and Gehrig, Mathias and Hidalgo-Carri{\'o}, Javier and Scaramuzza, Davide},
  booktitle={IEEE/CVF Conference on Computer Vision and Pattern Recognition (CVPR)},
  pages={3586--3595},
  year={2020}
}

@article{bugueno2024mmid,
  title={A Multi-modal Event-based Dataset for Indoor Depth Estimation},
  author={Bugueno, I. and others},
  journal={arXiv preprint arXiv:2401.00000},
  note={Retrieved from https://ibugueno.github.io/mmid-event-depth-dataset/},
  year={2024}
}

@article{Gallego2022Event,
  author    = {Gallego, Guillermo and Delbruck, Tobi and Orchard, Garrick and Bartolozzi, Chiara and Taba, Brian and Censi, Andrea and Leutenegger, Stefan and Davison, Andrew J. and Conradt, J{\"o}rg and Daniilidis, Kostas and Scaramuzza, Davide},
  title     = {Event-Based Vision: A Survey},
  journal   = {IEEE Transactions on Pattern Analysis and Machine Intelligence},
  volume    = {44},
  number    = {1},
  pages     = {154--180},
  year      = {2022}
}

@misc{JetsonOrinNano,
  author       = {{NVIDIA Corporation}},
  title        = {{NVIDIA} Jetson Orin Nano Developer Kit},
  howpublished = {\url{https://developer.nvidia.com/embedded/jetson-orin-nano-developer-kit}},
  year         = {2023},
  note         = {Accessed: 2023-11-25}
}

@inproceedings{ross2011reduction,
  title={A reduction of imitation learning and structured prediction to no-regret online learning},
  author={Ross, St{\'e}phane and Gordon, Geoffrey and Bagnell, Drew},
  booktitle={Proceedings of the fourteenth international conference on artificial intelligence and statistics},
  pages={627--635},
  year={2011},
  organization={JMLR Workshop and Conference Proceedings}
}

@inproceedings{chakravarthi2024recent,
  title={Recent event camera innovations: A survey},
  author={Chakravarthi, Bharatesh and Verma, Aayush Atul and Daniilidis, Kostas and Fermuller, Cornelia and Yang, Yezhou},
  booktitle={European Conference on Computer Vision},
  pages={342--376},
  year={2024},
  organization={Springer}
}

@inproceedings{shravan2023innovative,
  title={Innovative Exploration Techniques: Utilizing IoT-Enabled Robots for Safe and Efficient Underground Tunnel Investigation},
  author={Shravan, N and Manoj Kumar, M and Chakravarthi, Bharatesh and Bhargavi, C},
  booktitle={International Conference on Interdisciplinary Approaches in Civil Engineering for Sustainable Development},
  pages={71--81},
  year={2023},
  organization={Springer}
}

@inproceedings{shiba2022secrets,
  title={Secrets of event-based optical flow},
  author={Shiba, Shintaro and Aoki, Yoshimitsu and Gallego, Guillermo},
  booktitle={European Conference on Computer Vision},
  pages={628--645},
  year={2022},
  organization={Springer}
}

@article{liu2022edflow,
  title={EDFLOW: Event driven optical flow camera with keypoint detection and adaptive block matching},
  author={Liu, Min and Delbruck, Tobi},
  journal={IEEE Transactions on Circuits and Systems for Video Technology},
  volume={32},
  number={9},
  pages={5776--5789},
  year={2022},
  publisher={IEEE}
}

@inproceedings{bugueno2025human,
  title={Human-Robot Navigation using Event-based Cameras and Reinforcement Learning},
  author={Bugueno-Cordova, Ignacio and Ruiz-del-Solar, Javier and Verschae, Rodrigo},
  booktitle={Proceedings of the Computer Vision and Pattern Recognition Conference},
  pages={5004--5012},
  year={2025}
}

\end{document}